\documentclass[11pt]{article}

\usepackage[final]{acl}

\usepackage{times}
\usepackage{latexsym}

\usepackage[T1]{fontenc}

\usepackage[utf8]{inputenc}

\usepackage{microtype}

\usepackage{inconsolata}

\usepackage{graphicx}

\usepackage{subcaption}
\newcommand\sr[1]{\textbf{\textcolor{purple}{SR: #1}}}
\title{\textit{Which LoRA?} An Empirical Study on the Effectiveness of LoRA Techniques During Multilingual Instruction Tuning}

\author{Thamali Wijewardhana \and  Napoleon H.\ Reyes \and  Surangika Ranathunga\\
  School of Mathematical and Computational Sciences \\
  Massey University, \\
  Auckland, New Zealand \\
  \texttt{\{N.Wiewardhana, N.H.Reyes, S.Ranathunga\}@massey.ac.nz}}

\begin{document}
\maketitle
\begin{abstract}
We investigate whether commonly available LoRA variants have an advantage over basic LoRA in multilingual instruction tuning. Experiments involving LoRA and four other variants on two datasets across diverse target languages show that there is no significant advantage in using more complex LoRA variants instead of basic LoRA, with respect to balancing cross-lingual transfer and knowledge retention. An analysis of hidden embeddings reveal that layer-wise language representation remains largely similar across LLMs fine-tuned with different LoRA techniques, suggesting that architectural novelty of LoRA techniques may not translate into better cross-lingual adaptation.
\end{abstract}

\section{Introduction}


Low-Rank Adaptation (LoRA)~\citep{hu2022lora} is a prominent parameter-efficient fine-tuning (PEFT) technique that trains only a small set of low-rank matrices while freezing the original LLM parameters. Due to its success in getting LLMs adapted to new tasks and languages with fewer resources, many novel LoRA variants are being introduced \citep{yang2024low}. 
While they have been evaluated on English tasks, their efficacy is underexplored in the context of multilingual instruction tuning. 

With the introduction of multilingual LLMs, instruction tuning is being used to adapt them for tasks in new languages~\cite{garcia2026think}, and multilingual instruction tuning has shown to be better than its monolingual counterpart~\cite{shaham2024multilingual}. In multilingual instruction tuning, effective adaptation requires optimising multilingual transfer (the process of transferring knowledge learned in one language to another language) while preserving previously learned knowledge.  Whether recent LoRA variants are better suited in achieving these is an open question. 

Some research studied LoRA in cross-lingual settings, focusing on adapting them to low-resource languages (LRLs) \citep{khade2025challenges}, factors impacting cross-lingual transfer~\citep{khelli2025causes}, improving language learning and knowledge retention \citep{owodunni2025continually}, and empirical studies on data availability \citep{whitehouse2024low}. Furthermore, \citet{hassan2026grasp} proposed a method for effectively merging LoRA adapters for cross-lingual transfer. However, these investigations have been limited to basic LoRA.
 To the best of our knowledge, no study has comprehensively compared and analysed these LoRA variants in multilingual instruction tuning, particularly considering Low-resource languages (LRLs). 
While several recent studies have investigated the internal language representations of LLMs \citep{zhao2024large, wendler2024llamas, gurgurov2025language, tamo2026linguamap}, to the best of our knowledge, none have analyzed the impact of LoRA-based instruction tuning on these representations.

We conduct an empirical evaluation of a selected set of LoRA variants for their ability to facilitate cross-lingual transfer while preserving previously gained knowledge during multilingual instruction tuning. We mainly focus on low-resource settings, selecting five linguistically diverse languages, Urdu (Ur), Swahili (Sw), Hindi (Hi), Bengali (Bn) and Telugu (Te) (referred to as target languages (TLs)) (See Appendix \ref{app:languages} for language details). Our experiments reveal that more complex LoRA variants do not have a significant advantage over basic LoRA, indicating that architectural modifications introduced in recent LoRA variants do not always translate into performance gains in multilingual instruction tuning.
Using a layer-wise language representation analysis, we show that LoRA variants have not introduced noticeable language representation changes in the LLM. We also find that, in contrast to LoRA-based pre-training~\cite{owodunni2025continually}, in instruction tuning, LoRA should be applied to all the layers, instead of just the final layers.   Compared to \citet{imam2026full}, the closest to ours who focused only on African ASR, our work provides a broader investigation of language learning in LLMs (wider range of languages, model architectures, TL data mixtures, analyze the internal representations and layer-wise behavior of LLMs).

\section{Methodology}

\subsection{LoRA Variants}
Out of the numerous LoRA variants available~\citep{yang2024low}, we selected the following: Weight-Decomposed Low-Rank Adaptation (\textbf{DoRA}), Vector-Based Random Matrix Adaptation (\textbf{VeRA}), Adaptive Low-Rank Adaptation (\textbf{AdaLoRA}), Principal
Singular Values and Singular Vectors Adaptation (\textbf{PiSSA}). This selection was based on their level of acceptance by the research community\footnote{By considering the publication venue, number of citations for the relevant paper, their results as reported by other research.}, the availability of their open-source implementations, and their computational feasibility within our available GPU resources.


\textbf{LoRA} (baseline) optimises LLM training by approximating weight updates through low-rank decomposition \citep{hu2022lora}. \textbf{DoRA} decomposes pretrained  weights into two distinct components (magnitude and direction) and fine-tunes both~\citep{liu2024dora}. \textbf{VeRA} employs one pair of frozen, randomly initialised matrices that are shared across all adapted layers, and inserts trainable scaling vectors for layer-wise adaptation \citep{kopiczkovera}. \textbf{AdaLoRA} dynamically redistributes the parameter budget across the weight matrices based on their importance score~\citep{zhangadaptive}. \textbf{PiSSA} uses singular value decomposition to decompose the original weight matrix and trains only the principal components~\citep{meng2024pissa}. Appendix \ref{app:lora_techniques} has more details on these LoRA techniques. 

\subsection{Main Experiments}
Given that multilingual instruction tuning is better than monolingual instruction tuning~\cite{shaham2024multilingual} (we confirm the same; See Table~\ref{tab:data_compositions}), we fine-tune the LLM with a mixture of English and TL data, and measure the performance on both languages. As the first step, we determine the optimal data mixture percentage between English and the TL to be used during instruction tuning. since \citet{shaham2024multilingual} reported that during full parameter fine-tuning, even minimal multilingual exposure can significantly facilitate cross-lingual transfer, we evaluate four TL ratios: 0\%, 1\%, 10\%, and 50\% for LoRA-based fine-tuning.

Our investigation on the impact of hyper-parameters on different LoRA techniques follows a two-step approach. First, we assess each method using its optimal hyper-parameters. Second, to ensure a controlled comparison under a similar parameter budget, we fix the rank r to 8 and apply adapters to all linear layers of the transformer architecture. 

Note that all except DoRA converge to an optimal rank of 8 (parameter budget of 0.26\%) during hyperparameter tuning. DoRA reaches its peak performance at a r=16, with adapters applied to Q, K, V, Up and Down layers, resulting in a slightly larger parameter budget of 0.36\% trainable parameters. 
To ensure a fair comparison, we introduce DoRA*, in which r = 8 and adapters are applied to all linear layers, matching the parameter budget of others.
Table \ref{tab:param_budgets} presents the parameter budget and rank for each adapter configuration. The marginal increase
of 0.01\% in the number of trainable parameters for DoRA is due to the integration
of learnable magnitude components~\cite{liu2024dora}.
VeRA needs a higher rank under a similar parameter budget, which  is infeasible due to hardware constraints.

\begin{table}[t]
\tiny
  \centering
  \begin{tabular}{lcccccc}
    \hline
     & \textbf{LoRA} & \textbf{DoRA} & \textbf{DoRA*} & \textbf{VeRA} & \textbf{AdaLoRA} & \textbf{PiSSA} \\
    \hline
    
    Rank & 8 & 16 & 8 & 1024 & 8 & 8 \\
    \# Params(\%) & 0.26 & 0.36 & 0.27 & 0.02 & 0.26 & 0.26 \\ \hline
  \end{tabular}
    \caption{Rank and number of trainable parameters.}
  \label{tab:param_budgets}
\end{table}
  \begin{table*}
\tiny
  \centering
  \begin{tabular}{lccccccc}
    \hline
    Ln & \% & \textbf{LoRA} & \textbf{DoRA} &
    \textbf{DoRA*}  & \textbf{VeRA}  & \textbf{AdaLoRA}  & \textbf{PiSSA} \\ 
    \hline
    Ur & 0 & $\textbf{89.66}_{\pm 0.4}$/$\textnormal{65.36}_{\pm 1.7}$ & $\textnormal{89.22}_{\pm 1.2}$/$\textnormal{65.76}_{\pm 0.5}$ & $\textnormal{89.11}_{\pm 0.9}$/$\textnormal{66.16}_{\pm 1.5}$ & $\textnormal{86.85}_{\pm 0.5}$/$\textnormal{64.80}_{\pm 1.0}$ & $\textnormal{89.24}_{\pm 1.0}$/$\textnormal{66.49}_{\pm 1.4}$  & $\textnormal{88.61}_{\pm 0.7}$/$\textnormal{64.90}_{\pm 0.9}$ \\ 
    & 1 & $\textnormal{89.42}_{\pm 0.7}$/$\textbf{67.18}_{\pm 1.3}$ &  $\textnormal{88.98}_{\pm 0.4}$/$\textnormal{66.26}_{\pm 2.7}$ & $\smash{\underline{\textnormal{89.60}}}_{\pm 0.5}$/$\textnormal{65.92}_{\pm 2.9}$ & $\textnormal{86.80}_{\pm 0.1}$/$\textnormal{64.24}_{\pm 0.6}$ & $\textnormal{89.17}_{\pm 0.7}$/$\smash{\underline{\textnormal{66.52}}}_{\pm 1.6}$ & $\textnormal{88.49}_{\pm 0.7}$/$\textnormal{64.94}_{\pm 1.6}$ \\
    & 10 &  $\textnormal{88.72}_{\pm 0.6}$/$\textnormal{60.31}_{\pm 2.3}$ &  $\textnormal{88.77}_{\pm 1.0}$/$\textnormal{62.92}_{\pm 5.8}$ & $\textnormal{89.47}_{\pm 0.3}$/$\textnormal{65.89}_{\pm 2.9}$ & $\textnormal{87.10}_{\pm 0.1}$/$\textnormal{64.03}_{\pm 1.2}$ & $\textnormal{89.49}_{\pm 0.8}$/$\textnormal{58.92}_{\pm 1.8}$ & $\textnormal{87.70}_{\pm 0.4}$/$\textnormal{65.96}_{\pm 3.1}$ \\
    & 50 &  $\textnormal{87.16}_{\pm 0.4}$/$\textnormal{61.73}_{\pm 2.5}$ &  $\textnormal{86.97}_{\pm 0.5}$/$\textnormal{64.53}_{\pm 1.3}$ & $\textnormal{87.50}_{\pm 0.9}$/$\textnormal{64.01}_{\pm 2.9}$ & $\textnormal{84.80}_{\pm 0.3}$/$\textnormal{59.77}_{\pm 0.6}$ & $\textnormal{88.98}_{\pm 0.7}$/$\textnormal{61.99}_{\pm 2.4}$ & $\textnormal{86.28}_{\pm 1.6}$/$\textnormal{61.53}_{\pm 2.7}$ \\\hline
    Sw & 0 & $\textnormal{89.67}_{\pm 0.4}$/$\textnormal{61.10}_{\pm 2.8}$ &  $\textnormal{89.22}_{\pm 1.2}$/$\textnormal{62.30}_{\pm 0.5}$ & $\textnormal{89.11}_{\pm 0.9}$/$\textnormal{62.61}_{\pm 1.7}$ & $\textnormal{86.85}_{\pm 0.5}$/$\textnormal{59.65}_{\pm 1.7}$ & $\textnormal{89.24}_{\pm 0.3}$/$\textnormal{62.66}_{\pm 3.4}$ & $\textnormal{88.61}_{\pm 0.7}$/$\textnormal{62.22}_{\pm 1.7}$ \\
    & 1 & $\textnormal{89.61}_{\pm 0.2}$/$\textnormal{64.36}_{\pm 1.7}$ &  $\textnormal{89.38}_{\pm 0.1}$/$\textnormal{64.37}_{\pm 2.4}$ & $\smash{\underline{\textnormal{89.88}}}_{\pm 0.1}$/$\textbf{65.52}_{\pm 3.5}$ & $\textnormal{86.85}_{\pm 0.1}$/$\textnormal{62.61}_{\pm 1.6}$ & $\textbf{89.93}_{\pm 0.5}$/$\textnormal{64.56}_{\pm 1.2}$ & $\textnormal{88.75}_{\pm 0.3}$/$\textnormal{63.50}_{\pm 0.2}$ \\
    & 10 & $\textnormal{88.98}_{\pm 0.3}$/$\textnormal{59.27}_{\pm 2.7}$ &  $\textnormal{88.96}_{\pm 0.5}$/$\smash{\underline{\textnormal{65.09}}}_{\pm 0.9}$ & $\textnormal{89.22}_{\pm 0.3}$/$\textnormal{64.93}_{\pm 1.4}$ & $\textnormal{87.43}_{\pm 0.2}$/$\textnormal{63.54}_{\pm 1.8}$ & $\textnormal{88.19}_{\pm 0.5}$/$\textnormal{56.94}_{\pm 5.0}$ & $\textnormal{87.73}_{\pm 0.6}$/$\textnormal{64.02}_{\pm 0.9}$  \\
    & 50 & $\textnormal{87.40}_{\pm 1.2}$/$\textnormal{59.83}_{\pm 2.8}$ &  $\textnormal{87.02}_{\pm 1.2}$/$\textnormal{63.04}_{\pm 2.8}$ & $\textnormal{87.49}_{\pm 1.1}$/$\textnormal{62.71}_{\pm 2.9}$ & $\textnormal{85.26}_{\pm 0.7}$/$\textnormal{63.92}_{\pm 2.9}$ & $\textnormal{88.17}_{\pm 0.4}$/$\textnormal{60.20}_{\pm 2.0}$ & $\textnormal{87.77}_{\pm 0.2}$/$\textnormal{63.25}_{\pm 1.1}$  \\
    \hline
    Hi & 0 & $\textbf{89.67}_{\pm 0.4}$/$\textnormal{72.31}_{\pm 1.0}$ &  $\textnormal{89.22}_{\pm 1.2}$/$\textnormal{72.77}_{\pm 0.7}$ & $\textnormal{89.11}_{\pm 0.9}$/$\textnormal{73.56}_{\pm 1.4}$ & $\textnormal{86.85}_{\pm 0.5}$/$\textnormal{71.91}_{\pm 0.8}$ & $\textnormal{89.24}_{\pm 1.0}$/$\textnormal{73.81}_{\pm 1.5}$ & $\textnormal{88.61}_{\pm 0.7}$/$\textnormal{73.07}_{\pm 2.2}$  \\
    & 1 & $\textnormal{89.31}_{\pm 0.4}$/$\textnormal{74.81}_{\pm 1.5}$ &  $\textnormal{89.37}_{\pm 0.6}$/$\textbf{75.86}_{\pm 0.8}$ & $\smash{\underline{\textnormal{89.66}}}_{\pm 0.5}$/$\textnormal{74.47}_{\pm 1.8}$ & $\textnormal{86.81}_{\pm 0.1}$/$\textnormal{73.00}_{\pm 0.3}$ & $\textnormal{88.49}_{\pm 1.0}$/$\textnormal{72.71}_{\pm 2.5}$ & $\textnormal{88.28}_{\pm 0.6}$/$\textnormal{73.40}_{\pm 0.4}$  \\
    & 10 & $\textnormal{88.99}_{\pm 0.6}$/$\textnormal{73.77}_{\pm 2.3}$ &  $\textnormal{89.26}_{\pm 0.1}$/$\textnormal{75.59}_{\pm 0.2}$ & $\textnormal{89.35}_{\pm 0.7}$/$\textnormal{74.80}_{\pm 1.2}$ & $\textnormal{87.03}_{\pm 0.2}$/$\textnormal{70.36}_{\pm 5.9}$ & $\textnormal{89.63}_{\pm 0.6}$/$\smash{\underline{\textnormal{75.49}}}_{\pm 0.7}$ & $\textnormal{88.13}_{\pm 0.2}$/$\textnormal{75.44}_{\pm 0.5}$  \\ 
    & 50 & $\textnormal{87.87}_{\pm 1.8}$/$\textnormal{73.71}_{\pm 1.4}$ &  $\textnormal{88.01}_{\pm 0.7}$/$\textnormal{74.68}_{\pm 0.5}$ & $\textnormal{87.24}_{\pm 1.7}$/$\textnormal{73.04}_{\pm 1.9}$ & $\textnormal{85.55}_{\pm 0.7}$/$\textnormal{73.38}_{\pm 0.9}$ & $\textnormal{88.84}_{\pm 0.5}$/$\textnormal{73.68}_{\pm 1.5}$ & $\textnormal{87.31}_{\pm 0.4}$/$\textnormal{74.27}_{\pm 1.6}$  \\
    \hline
  \end{tabular}

  \caption{ F1 scores across LoRA techniques during Llama-3.1-8B fine-tuning on XNLI (averaged over 3 random seeds). Each value is reported as \textbf{English F1 / Target-language F1}. \textbf{Bold} indicates the best result for each language and \underline{underline} the second-best. Ln: Language; \%: LRL percentage.}
  
  \label{tab:performance_optimal_parameters}
\end{table*} 

In order to verify the  generalizability of our findings, we carried out a series of ablation studies with respect to the size and the architecture of the LLM changes, when the task changes, and when the LoRA rank changes.




\section{Experimental Setup}

\textbf{Data: }We use the cross-lingual natural language inference (XNLI) dataset~\citep{conneau2018xnli} for main experiments (with Ur, Sw, Hi,), and TyDiQA-GoldP \citep{clark2020tydi}, a question answering dataset for the ablation (with Sw, Bn, Te). 
For XNLI, we use 5,000 samples for training, 500 for validation and the full test portion (5010 samples) for testing. 
For TyDiQA-GoldP, we use subsets of the training set for training and validation (3000 for training, 696 for validation) and the entire validation set for testing (Following \citet{ahuja2023mega}).
For both datasets, we apply stratified random sampling. 
We use micro F1 score as evaluation metric.


\textbf{Models: } Preliminary experiments (Table~\ref{tab:data_compositions}) show that Llama-3.1-8B base is better than its instruct version. Therefore the former is used for subsequent experiments. To ensure generalizability of our findings, we extend a subset of our experiments to Llama-3.2-3B base (same
model family, different size) and Qwen3-8B base (different model family, same size). We conduct a hyperparameter search starting from the values recommended for each LoRA variant (see Appendix \ref{app:hyperparameters} and Appendix \ref{app:infrastructure}. 
\textbf{Our code will be released}. 


\section{Results}



\begin{table}
\tiny
  \centering
  \begin{tabular}{lcccccc}
    \hline
    \textbf{Metric} & \textbf{LoRA} & \textbf{DoRA} & \textbf{DoRA*} & \textbf{VeRA} & \textbf{AdaLoRA} & \textbf{PiSSA} \\
    \hline
    ATT (Min) & 46.12 & 88.14 & 110.25 & 45.83 & \textbf{44.63} & 76.75 \\
    AGU (\%) & \textbf{37.15} & 84.91 & 85.98 & 71.71 & 46.39 &  42.25 \\
    AMU (Mib) & 22373 & 32633 & 38801 & 27478 & 21947 & \textbf{21250} \\
    MMU (Mib) & 23387 & 33077 & 39281 & 28885 & 23571  & \textbf{22087} \\
    AGT ($^\circ$C) & \textbf{41.76} & 63.71 & 61.43 & 64.41 & 56.93 & 51.51 \\ \hline
  \end{tabular}
  \caption{Comparison of computational efficiency metrics across LoRA techniques during Llama-3.1-8B fine-tuning on XNLI. ATT: Avg. Training Time; AGU: Avg. GPU Utilisation; AMU: Avg. Memory Usage; MMU: Max. Memory Usage; AGT: Avg. GPU Temperature. }
  \label{tab:resource_usage_optimal_parameters}
\end{table}

\subsection{Main Experiments}

Table \ref{tab:performance_optimal_parameters} summarises the results (across 3 seeds) of instruction tuning Llama-3.1-8B with XNLI. Similar to findings for full-fine-tuning~\cite{shaham2024multilingual},  introduction of even a small proportion (1\%) of TL data enhances performance in the TLs across configurations, with only three exceptions, and just a slight decline in En performance in most cases. 
Increasing the TL proportion further (to 10\% and 50\%) results in performance degradation for both English and TLs in most configurations. 

Considering the peak performance across all LoRA types and TL percentages, the highest overall F1 scores for Ur, Sw, and Hi are achieved by LoRA, DoRA*, and DoRA, respectively. However, an ANOVA test performed across all four language percentages indicates no statistically significant difference ($F = 0.33$, $p = 0.89$) among the LoRA variants on TL performance.
On En performance, there is a significant difference($F = 5.14$, $p = 0.0002$) and post-hoc comparisons via Tukey's HSD yield a distinct performance hierarchy: LoRA = DoRA = DoRA* = AdaLoRA > PiSSA > VeRA. VeRA forces all layers to share identical global $A$ and $B$ matrices, restricting updates to a single subspace. We believe that this shared configuration is a potential disadvantage for prior knowledge preservation. 


As reported in Table \ref{tab:resource_usage_optimal_parameters}, no single LoRA variant dominates across all efficiency dimensions. LoRA demonstrates a favorable efficiency profile, with AdaLoRA being the second best. DoRA and DoRA* demonstrate the least favorable efficiency profile (ranked at the bottom for 4/5 of the metrics). 


\subsection{Experimenting with Different LLMs} \label{subsec: different_LLMs}

\begin{table}
\tiny
  \centering
  \begin{tabular}{lccccc}
    \hline
      & \textbf{LoRA} & \textbf{DoRA}  & \textbf{VeRA}  & \textbf{AdaLoRA}  & \textbf{PiSSA}  \\ 
    \hline
    L & \underline{86.96}/63.13 & \textbf{87.30}/\underline{63.25} & 83.82/59.41 & \underline{86.96}/\textbf{64.04} &  85.36/62.07\\ 
    Q & \textbf{91.13}/\textbf{64.34} & 90.70/\underline{63.97} & 89.50/62.73 & 90.71 / 63.41 & \underline{90.75}/63.74 \\ 
    \hline
  \end{tabular}

   \caption{Average F1 scores of LoRA techniques on the XNLI dataset for Llama-3.2-3B (L) and Qwen3-8B (Q).}
   
  \label{tab:different_LLMs}
\end{table} 

We use 1\% ratio (which gave the best result) in subsequent experiments. Table \ref{tab:different_LLMs} illustrates the average performance across all three TLs for both Llama-3.2-3B and Qwen-3-8B. Detailed results are in Table \ref{tab:performance_different_model_size} and Table \ref{tab:performance_different_model} in Appendix \ref{app:extended_results}. 
Average F1 scores, and statistical testing (ANOVA and post-hoc) indicate that VeRA significantly ($p < 0.05$) underperforms other methods across both TL and English in the Llama-3.2-3B experiments. For Qwen 3-8B ANOVA testing indicates no statistically significant difference ($p = 0.65$) among the LoRA methods on TL performance. For English, VeRA significantly underperforms ($p < 0.05$) other methods.  

\subsection{Experimenting with Varying Ranks} \label{subsec: vary_param_budget}

\begin{table}
\tiny
  \centering
  \begin{tabular}{lcccc}
    \hline
     Rank & \textbf{LoRA} & \textbf{DoRA*}  & \textbf{AdaLoRA}  & \textbf{PiSSA}  \\ 
    \hline
    8 & \underline{89.47}/\textbf{67.87} & \textbf{89.52}/\underline{67.19} & 89.15/65.71 & 89.39/66.26\\ 
    16 & \underline{89.18}/\underline{64.15} & \textbf{89.75}/\textbf{65.23} & 88.39/61.86 & 88.41/63.12  \\ 
    32 & \textbf{89.82}/\underline{68.76} & \underline{89.62}/\textbf{69.01} & 89.14/68.30 & 88.22/66.86\\ 
    \hline
  \end{tabular}

   \caption{Average F1 of LoRA techniques on Llama-3.1-8B across different ranks for the XNLI dataset.}
    \label{tab:different_ranks}
\end{table}

To investigate the sensitivity of each adapter to the parameter budget, we evaluate performance across rank values. Here, the parameter budget is directly proportional to the rank as we keep other factors constant. For these experiments, we use DoRA*, which maintains a parameter budget comparable to the other methods. As per Table \ref{tab:different_ranks} (detailed results in Table \ref{tab:performance_varying_param_budget} in Appendix \ref{app:extended_results}), TL performance exhibits sharp sensitivity to rank scaling, while En F1 scores remaining highly stable across all four techniques. TL performance uniformly degrades at rank 16 relative to Rank 8, while rank 32 yields peak TL performance across nearly all variants.



\subsection{Evaluating Robustness on TyDiQA}

As per Table \ref{tab:performance_different_tydiqa}, across all 3 languages, performance differences among the LoRA variants on Llama-3.1-8B 
are minor ($p = 0.30$ for TLs and $p = 0.10$ for English), validating our earlier observations. 

\begin{table}
\tiny
  \centering
  \begin{tabular}{l@{\hspace{6pt}}ccccc}
    \hline
     & \textbf{LoRA} & \textbf{DoRA}  & \textbf{VeRA}  & \textbf{AdaLoRA}  & \textbf{PiSSA}  \\ 
    \hline
    Sw & \textbf{83.49}/\textbf{82.94} & 82.34/82.38 & \underline{82.64}/80.92 & 82.34/\underline{82.77} & 82.37/81.34 \\ 
    Bn & \textbf{84.40}/55.01 & 82.92/56.35 & 82.13/\underline{56.65} & 80.64/50.34 & \underline{83.10}/\textbf{57.33} \\ 
    Te & \underline{82.60}/\underline{70.20} & 82.04/69.48 & 81.54/69.12 & 81.68/62.42 & \textbf{84.16}/\textbf{70.80}\\ 
    \hline
  \end{tabular}

   \caption{Results of LoRA techniques on the TyDiQA-GoldP for Llama 3.1-8B.}
   
  \label{tab:performance_different_tydiqa}
\end{table} 

\begin{figure}
\includegraphics[width=\columnwidth]{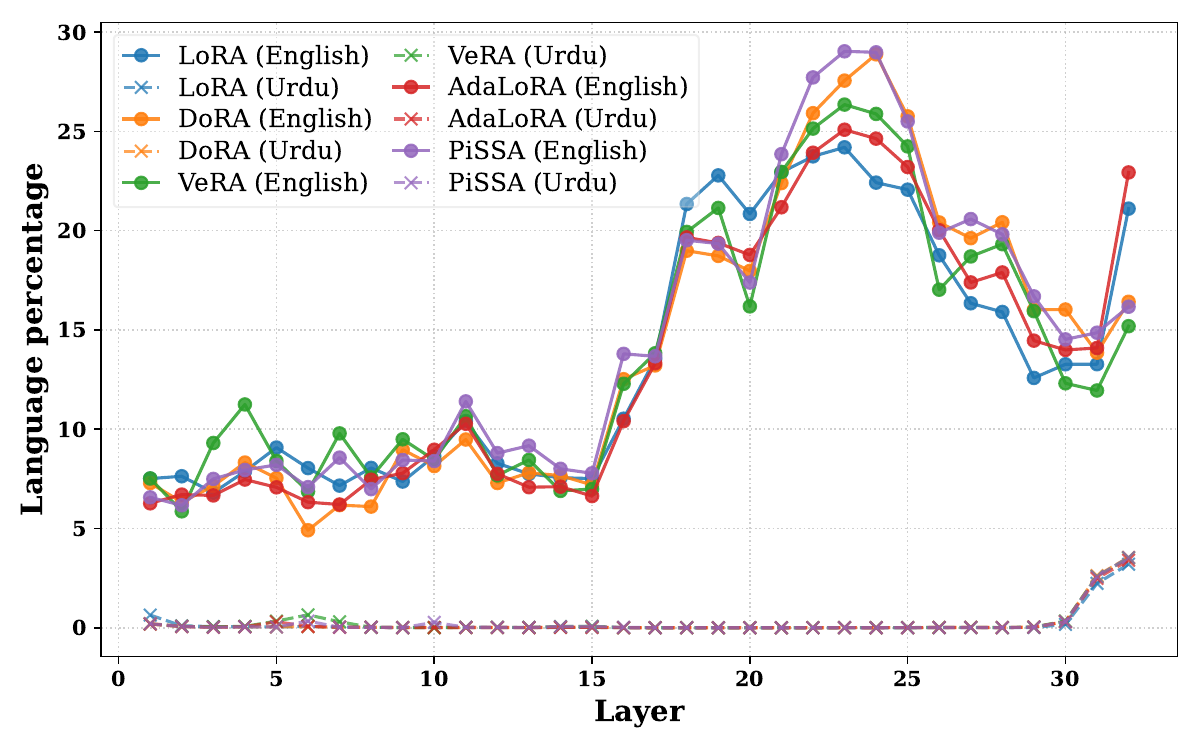}
  \caption{Layer-wise language distribution.}
  \label{fig:Urdu_layerwise}
\end{figure}

\subsection{Layer-wise Hidden Embedding Analysis}
Using the method proposed by~\citet{zhao2024large}, we perform a layer-wise analysis of embeddings of LLMs to investigate whether the LoRA variants have been able to introduce significant  language representation changes in the LLM.

Experiments were conducted on Ur, Sw, and Hi using Llama-3.1-8B fine-tuned with each LoRA variant with 1\% TL data. 
For each test instance, we classified decoded layer-wise embeddings into \textit{English}, \textit{TL}, or \textit{other} using CLD3\footnote{\url{https://pypi.org/project/pycld3/}}, computed per-layer language proportions, and averaged them over the test set. 
The percentages of En and Ur in the hidden embeddings of each layer are illustrated in Figure \ref{fig:Urdu_layerwise}. Experimental setup and supplementary results for additional languages are in Appendix \ref{app:layerwise_embedding_analysis}.

We note that the amount of language representation across layers is largely consistent across all LoRA variants, indicating that LoRA variants have not been able to introduce noticeable language representation changes in the LLM. 
These language representations exhibit a distinct three-stage pattern transitioning from initial layers, to English-dominant middle layers, and finally to outer layers where English dominance declines. This pattern aligns with findings of prior work that investigated language representations of the LLMs. ~\cite{zhao2024large,wendler2024llamas,tamo2026linguamap,gurgurov2025language}. 

TL percentages in the final layers (30-31), as well as the English percentages in the task-solving layers (22–27), both exhibit a highly significant ($p < 0.05$) positive Pearson correlation with the final TL F1 score (detailed results in Figure \ref{fig:correlation_heatmap} of Appendix \ref{app:layerwise_embedding_analysis}.). 
Motivated by this, we applied LoRA adapters exclusively to these layers. However, it under-performs compared to LoRA fine-tuning of all layers (see Table \ref{tab:selective_tune} of Appendix \ref{app:layerwise_embedding_analysis}). This contradicts with \citet{owodunni2025continually}, who reported that applying LoRA to the first ten and last two layers is sufficient during continual pretraining. 

\section{Conclusion}
Our comprehensive experiments reveal no significant difference among the considered LoRA techniques in TL learning, while VeRA underperformed the other LoRA techniques in terms of English language preservation.  
Layer-wise embedding and correlation analyses show that every LoRA technique, regardless of its underlying architecture converges on an identical "Reason in English, Translate at Exit" setup. Future work may focus on developing LoRA architectures that formalize and optimize this structural mechanic.


\section{Limitations}

While our evaluation encompasses a significant range of configurations, several constraints bound the scope of this study. First, due to computational resource limitations, we restricted our analysis to LLMs in the 3B to 8B parameter range. Future research could investigate whether the above observations persist in much larger models.

Our experimental evaluation is limited to two downstream tasks (NLI and QA). This restriction stems from computational constraints as well as the limited availability of task-specific fine-tuning datasets that simultaneously support LRLs, are feasible within our GPU budget, and cover tasks (i.e.~text classification and question answering) beyond those considered in this study. If the availability of LRL datasets spanning diverse tasks increases, future work could extend this analysis to a broader range of tasks.

Due to lack of computing resources, we could only experiment with 0\%, 1\%, 10\%, 50\% data mixtures. While future experiments can conduct a a denser sweep around 0–10\%, we believe this is not necessary for the current research, as our primary goal is not to identify the optimal data mixture. Similarly, we only checked the forgetting only in the context of English, because English is the language that is most prominent in LLMs and most language resources are available for English. 

Finally, while there many LoRA variants available, we only experiment with LoRA techniques that are supported by the Huggingface library and are feasible with our GPU resources.

\section{Ethical Considerations}
We use publicly available datasets. XNLI is under CC-BY-NC-4.0. TyDiQA-GoldP is under Apache 2.0. Whatever the biases in these datasets may have been reflected in our results.
\bibliography{custom}

\appendix


\begin{table*}[t]
  \centering
  \begin{tabular}{lccccc}
    \hline
    \textbf{Language} & \textbf{LoRA} & \textbf{DoRA}  & \textbf{VeRA}  & \textbf{AdaLoRA}  & \textbf{PiSSA}  \\ 
    \hline
    Urdu & 87.34/61.99 & \underline{87.50}/\textbf{63.41} & 83.95/58.10 & \textbf{87.82}/\underline{62.19} & 85.54/61.43 \\ 
    Swahili & \underline{86.72}/56.00 & \textbf{87.26}/\underline{57.24} & 83.81/53.75 & 86.40/\textbf{58.94} & 85.04/54.65 \\
    Hindi & \underline{86.82}/\textbf{71.41} & \textbf{87.16}/69.10 & 83.71/66.40 & 86.66/\underline{70.99} & 85.50/70.15 \\ 
    \hline
  \end{tabular}

\caption{F1 scores of LoRA techniques on the XNLI dataset for Llama-3.2-3B. Each value is reported as \textbf{English F1 / Target-language F1}. \textbf{Bold} indicates the best result for each language and \underline{underline} the second-best.}
  
  \label{tab:performance_different_model_size}
\end{table*} 

\begin{table*}
  \centering
  \begin{tabular}{lccccc}
    \hline
    \textbf{Language} & \textbf{LoRA} & \textbf{DoRA}  & \textbf{VeRA}  & \textbf{AdaLoRA}  & \textbf{PiSSA}  \\ 
    \hline
    Urdu & \underline{91.21}/\underline{72.21} & \textbf{91.31}/\underline{71.57} & 89.48/68.78 & 91.03/\textbf{72.35} & 90.69/70.63 \\ 
    Swahili & 90.65/60.09 & \underline{90.73}/60.35 & 89.62/\underline{61.31} & 90.57/58.52 & \textbf{90.77}/\textbf{61.63} \\
    Hindi & \textbf{91.53}/\textbf{60.73} & 90.07/\underline{60.00} & 89.40/58.12 & 90.53/59.36 & \underline{90.81}/58.98 \\ 
    \hline
  \end{tabular}
\caption{F1 scores of LoRA techniques on the XNLI dataset for Qwen3-8B. Each value is reported as \textbf{English F1 / Target-language F1}. \textbf{Bold} indicates the best result for each language and \underline{underline} the second-best.}
  \label{tab:performance_different_model}
\end{table*} 

\begin{table*}
  \centering
  \begin{tabular}{lcccc}
    \hline
    \textbf{Rank} & \textbf{LoRA} & \textbf{DoRA*} & \textbf{AdaLoRA}  & \textbf{PiSSA}  \\ 
    \hline
     & \multicolumn{3}{c}{\textbf{Urdu}} \\ \hline
    8 & \underline{89.42}/\textbf{68.64} & \textbf{89.88}/66.30 & 88.88/64.63 & \underline{88.82}/\underline{66.54} \\
    16 & \underline{89.26}/67.20 & 89.22/\textbf{67.64} & 89.16/64.77 & \textbf{90.05}/\underline{67.52} \\ 
    32 & \textbf{89.72}/\textbf{67.76} & \underline{89.46}/67.64 & 89.42/\underline{67.74} & 89.30/64.71 \\ \hline
    & \multicolumn{3}{c}{\textbf{Swahili}} \\ \hline
    8 & \underline{89.64}/\underline{64.95} & \textbf{90.05}/\textbf{67.24} & 89.34/63.67 & 88.74/63.49 \\ 
    16 & 88.18/63.19 & \underline{89.32}/\textbf{64.43} & \textbf{89.48}/60.27 & 89.06/\underline{63.49} \\
    32 & \underline{89.72}/\underline{64.31} & 89.88/64.03 & \textbf{89.34}/\textbf{61.65} & 87.42/62.37 \\ 
    \hline
    & \multicolumn{3}{c}{\textbf{Hindi}} \\ \hline
    8 & \underline{89.78}/\textbf{76.60} & \textbf{89.80}/74.43 & 89.50/\underline{75.54} & 88.64/73.77 \\ 
    16 & \textbf{90.07}/\underline{76.16} & \underline{89.65}/\textbf{76.24} & 89.38/74.37 & 89.54/74.29 \\ 
    32 & \textbf{90.03}/\underline{74.21} & \underline{89.54}/\textbf{75.36} & 88.70/72.51 & 87.96/73.51 \\ 
    \hline
  \end{tabular}
\caption{
	F1 scores of LoRA techniques on XNLI dataset across different ranks using a 1\% TL data ratio. Each value is reported as \textbf{English F1 / Target-language F1}. \textbf{Bold} indicates the best result for each rank and \underline{underline} the second-best.}
  
  \label{tab:performance_varying_param_budget}
\end{table*} 




\section{Languages} \label{app:languages}

For XNLI, we conducted our experiments on three distinct languages: Urdu, Swahili, and Hindi. Urdu is a low-resource Indo-Aryan language written in the Perso-Arabic script, and shares significant linguistic similarities with Hindi, which is represented in pretraining corpora of the LLMs used in this study. Swahili is a low-resource Bantu language written in the Latin script \citep{owodunni2025continually}. Hindi is a medium-resource Indo-Aryan language written in the Devanagari script. For TyDiQA, we evaluate on Swahili as well as Bengali and Telugu, the latter two representing the Indo-Aryan and Dravidian language families and written in Bengali and Telugu scripts, respectively. The selection of languages with diverse scripts, language families and structures ensures the robustness of our experiments.

\section{LoRA Techniques} \label{app:lora_techniques}

This section provides a brief overview of the four LoRA variants used in our experiments.

\begin{itemize}
    \item \textbf{DoRA} is designed to bridge the performance gap between LoRA and full-parameter fine-tuning. It operates by decomposing the pre-trained weight matrices into magnitude and directional components. During the adaptation process, the directional component is updated via low-rank matrices while the magnitude is tuned independently. This method provides a learning ability closer to full fine-tuning without introducing any inference latency compared to LoRA.
    \item \textbf{VeRA} departs from traditional LoRA by utilising a single pair of random, low-rank matrices that remain frozen and are shared across all adapted layers. Layer-wise tuning is facilitated through scalable training vectors. This design achieves a substantial reduction of trainable parameters compared to LoRA, 
    while maintaining performance levels competitive with more parameter-intensive adaptation methods.
    \item \textbf{AdaLoRA} improves the efficiency of LoRA by moving away from a static parameter distribution across weight matrices. AdaLoRA utilises a Singular Value Decomposition (SVD) based formulation to represent weight increments, allowing the system to quantify the relative importance of specific updates. This technique prunes singular values with low importance scores, effectively reallocating the training budget toward incremental matrices that are more critical for the target task. This adaptive approach ensures that the model capacity is utilised where it provides the greatest performance gain.
    \item \textbf{PiSSA} targets the original weight matrix $W$ directly, in contrast to LoRA and its variants, which approximate the weight update $\Delta W$.
    In this approach, the weight matrix $W$ undergoes singular value decomposition (SVD), allowing it to be segmented based on the magnitude of its singular values. This process yields two components: a principal low-rank matrix, which captures the most significant singular values, and a residual matrix, containing the remaining smaller singular values. 
    During training, the principal component is updated, and the residual component remains fixed. 
\end{itemize}

\section{Extended Results}
\label{app:extended_results}

We provide the full per-language breakdown for the generalizability experiments discussed in Section \ref{subsec: different_LLMs} in Table \ref{tab:performance_different_model_size} and Table \ref{tab:performance_different_model}. Table \ref{tab:performance_different_model_size}  presents the F1 scores for Llama-3.2-3B, and Table \ref{tab:performance_different_model} details the results for Qwen 3-8B. All reported scores are derived from experiments using the 1\% TL data ratio.

Table \ref{tab:performance_varying_param_budget} presents the extended results discussed in Section \ref{subsec: vary_param_budget}. For each language and rank setting, we report the F1 scores for both English and the corresponding TL.

\section{Hyperparameters} \label{app:hyperparameters}

We initialized hyperparameters using the values recommended in the original method papers and subsequently performed controlled tuning to ensure a fair comparison across methods. 

We tuned the number of training epochs, learning rate, LoRA rank ($r$), LoRA alpha ($\alpha$) and target modules. Following the original design of VeRA, we explored larger rank values  $r \in \{256, 512, 1024, 2048\}$. For all other PEFT methods, we evaluated $r \in \{4, 8, 16, 32\}$ and  $\alpha \in \{4, 8, 16, 32\}$. Learning rates were swept across $\{2\times10^{-5}, 5\times10^{-5}, 1\times10^{-4}, 2\times10^{-4}, 5\times10^{-4}, 1\times10^{-3}, 2\times10^{-3}, 4\times10^{-3}\}$. 
The default number of training epochs was set to 3. However, if the original method papers recommended an alternative epoch count, we evaluated that configuration as well and compared it against the default baseline.
For target modules, we evaluated both the module configuration recommended in the respective original papers and an "all-linear" configuration, selecting the best-performing option for each method.
Hyperparameter configurations for the main experiments on the XNLI and TyDiQA-GoldP datasets are provided in Table \ref{tab:params_XNLI} and Table \ref{tab:params_TyDiQA}, respectively.
 For the ablation study conducted on XNLI with increased ranks $r \in \{16, 32\}$, we tuned the LoRA alpha over the set $\{16, 32, 64\}$ while keeping other hyperparameters consistent with the main configuration. Table \ref{tab:params_varying_rank}  summarizes the optimal $\alpha$ values chosen for each method across these increased ranks.

 \begin{figure}
\includegraphics[width=\columnwidth]{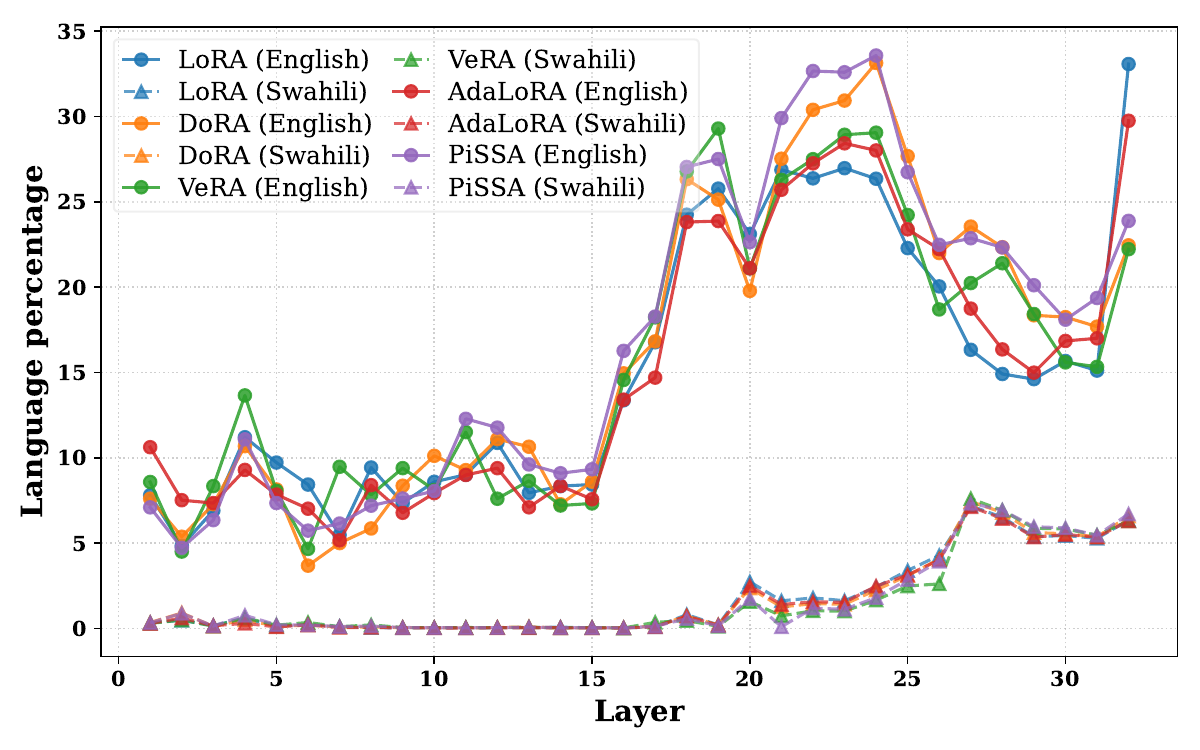}
  \caption{Layer-wise language distribution in hidden
embeddings for Sw experiments}
  \label{fig:swahili_layerwise}
\end{figure}

\begin{figure}
\includegraphics[width=\columnwidth]{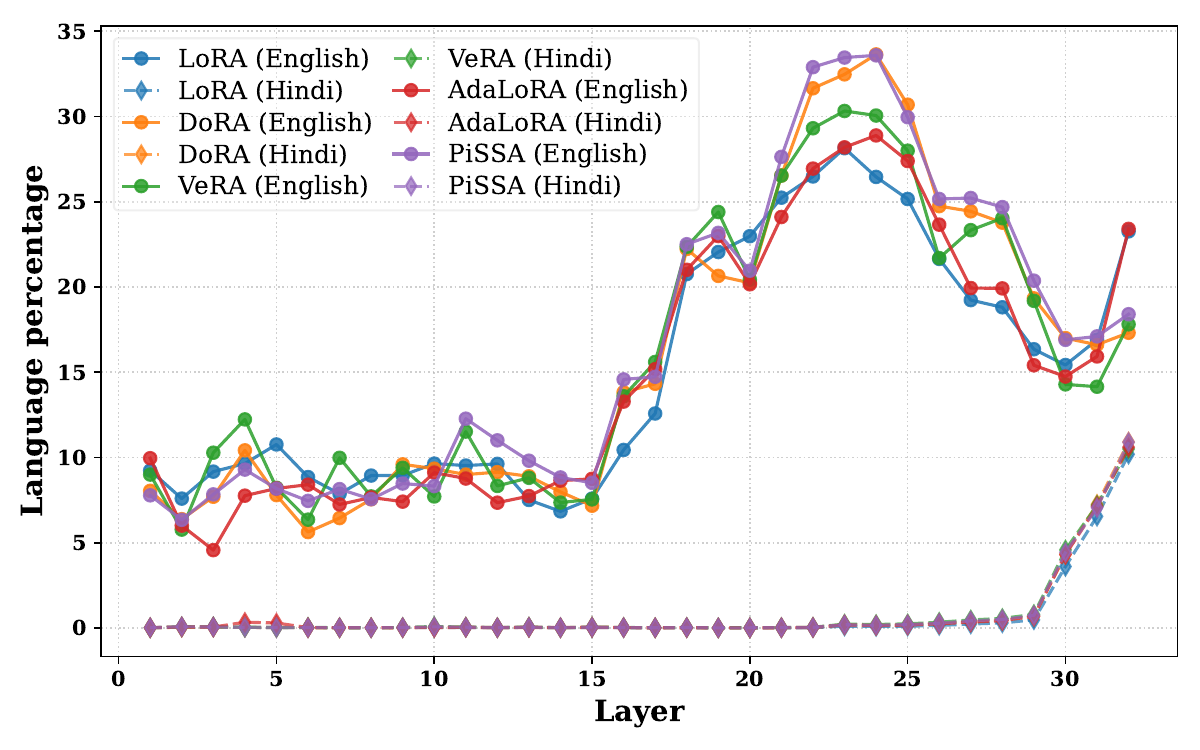}
  \caption{Layer-wise language distribution in hidden
embeddings for Hi experiments}
  \label{fig:hindi_layerwise}
\end{figure}

\begin{table*}
  \centering
  \begin{tabular}{lcccccc}
    \hline
    \textbf{Hyperparameter} & \textbf{LoRA} & \textbf{DoRA} & \textbf{DoRA*} & \textbf{VeRA} & \textbf{AdaLoRA} & \textbf{PiSSA} \\
    \hline
Rank & 8 & 16 & 8 & 1024 & 8 & 8 \\
Initial Rank & – & – & – &– & 12 & – \\
Alpha & 32 & 32 & 32 & – & 32 & 8 \\
Learning Rate & 2e-4 & 1e-4 & 1e-4 & 4e-3 & 2e-3 & 2e-5 \\
Epochs & 3 & 3 & 3 & 3 & 3 & 5 \\
Batch Size & 16 & 16 & 16 & 16 & 16 & 16 \\
Weight Decay & 0.01 & – & – & – & 0.01 & – \\
Warmup Ratio & 0.05 & – & – & 0.1 & 0.05 & 0.03 \\
Warmup Steps & – & 100 & 100 & – & – & – \\
Scheduler & Linear & Linear & Linear & Cosine & Linear & Cosine \\
Optimizer & AdamW & AdamW & AdamW & AdamW & AdamW & AdamW \\
Target Modules & all-linear & q,k,v,up,down & all-linear & all-linear & all-linear & all-linear \\
    \hline
  \end{tabular}
  \caption{\label{tab:params_XNLI}
    Hyperparameter configurations used for different LoRA methods in the main experiments.
  }
\end{table*}

\begin{table*}
  \centering
  \begin{tabular}{lccccc}
    \hline
    \textbf{Hyperparameter} & \textbf{LoRA} & \textbf{DoRA} & \textbf{VeRA} & \textbf{AdaLoRA} & \textbf{PiSSA} \\
    \hline
    Rank & 8 & 16 & 1024 & 8 & 8 \\
    Initial Rank & – & – & – & 12 & – \\
    Alpha & 8 & 16 & – & 32 & 8 \\
    Learning Rate & 5e-5 & 1e-4 & 2e-3 & 1e-3 & 2e-5 \\
    Epochs & 3 & 3 & 3 & 3 & 5 \\
    Batch Size & 16 & 16 & 16 & 16 & 16 \\
    Weight Decay & 0.01 & – & – & 0.01 & – \\
    Warmup Ratio & 0.05 & – & 0.1 & 0.05 & 0.03 \\
    Warmup Steps & – & 100 & – & – & – \\
    Scheduler & Linear & Linear & Cosine & Linear & Cosine \\
    Optimizer & AdamW & AdamW & AdamW & AdamW & AdamW \\
    Target Modules & all-linear & q,k,v,up,down & all-linear & all-linear & all-linear \\
    \hline
  \end{tabular}
  \caption{\label{tab:params_TyDiQA}
    Hyperparameters for LoRA methods on TyDiQA-GoldP.
  }
\end{table*}

\begin{table*}
  \centering
  \begin{tabular}{lcccc}
    \hline
    \textbf{Rank} & \textbf{LoRA} & \textbf{DoRA*} & \textbf{AdaLoRA}  & \textbf{PiSSA}  \\ 
    \hline
    16 & 16 & 16 & 16 & 16 \\ 
    32 & 32 & 32 & 32 & 32 \\
    \hline
  \end{tabular}
\caption{
	Best-performing alpha values for the XNLI rank sensitivity study}
  
  \label{tab:params_varying_rank}
\end{table*} 

\begin{table*}
  \centering
  \begin{tabular}{lcccc}
    \hline
    \textbf{Layers tuned} & \textbf{Ur} & \textbf{Sw} & \textbf{Hi}  \\ 
    \hline
    22-27 & 66.78/51.41 &  66.96/49.90 & 67.25/55.30\\ 
    30-31 & 66.08/45.55 &  66.62/48.96 & 65.94/51.91 \\ 
    22-31 & 68.82/51.73 & 68.72/50.89 & 69.26/55.54\\ 
    \hline
  \end{tabular}  
\caption{
	F1 scores on XNLI obtained by multilingual instruction-tuning selected layers of Llama-3.1-8B using LoRA. Each value is reported as \textbf{English F1 / Target-language F1}.}
\label{tab:selective_tune}
\end{table*} 

\begin{table*}
  \centering
  \begin{tabular}{lcccc}
    \hline
    & \multicolumn{2}{c}{\textbf{Base}} & \multicolumn{2}{c}{\textbf{Instruct}} \\
    \textbf{Tuning approach} & \textbf{2000} & \textbf{20000} & \textbf{2000} & \textbf{20000}  \\ 
    \hline
    Tune with English & 84.35/62.69 &  89.70/65.41 &85.12/56.46 & 90.41/65.26\\ 
    Tune with Ur &  58.72/52.29 & 74.45/63.73 & 68.38/56.48 & 82.67/64.29\\ 
    Tune with English + Ur & 86.54/65.24 &  90.21/65.52 & 86.14/62.09 & 90.63/63.07\\ 
    \hline
  \end{tabular}  
\caption{
	F1 scores on XNLI obtained by instruction-tuning Llama-3.1-8B base LLM and Llama-3.1-8B-Instruct with LoRA under varying instruction-tuning data compositions and dataset sizes (2,000 vs. 20,000). Each value is reported as \textbf{English F1 / Ur F1}. }
\label{tab:data_compositions}
\end{table*}

\section{Computing Infrastructure} \label{app:infrastructure}
All experiments were conducted using Pytorch \citep{paszke2019pytorch} and the HuggingFace
library \citep{wolf2020transformers} on an NVIDIA A100 GPU with 80GB memory. We used PyTorch version 2.5.1 and CUDA 12.2.

\section{Layer-wise Hidden Embedding Analysis}
\label{app:layerwise_embedding_analysis}

We conducted a layer-wise embedding analysis on Urdu, Swahili, and Hindi to track how internal language representations evolve across different fine-tuning techniques. For this analysis, we utilized the Llama-3.1-8B models fine-tuned on a training mixture consisting of 1\% TL data and 99\% English data. For each model, we passed the target-language test set and extracted the hidden state representations from the output of every transformer layer.
The extracted embeddings were decoded using the model’s language modeling head and 
classified into three language categories: English, TL, and other. Embeddings identified as
neither English nor the corresponding TL were assigned to the other category.

Language identification was performed using CLD3 library. For each test instance and each layer, we computed the proportion of embeddings assigned to each language category. These layer-wise language proportions were then averaged across the entire test set to obtain the final distributional patterns.

To ensure robustness of our findings, we repeated a subset of the analysis using the Lingua \footnote{\url{https://pypi.org/project/lingua/}}language identification library. The resulting layer-wise language distribution patterns were consistent with those obtained using CLD3, confirming that our observations are not dependent on a specific language identification tool. Figure \ref{fig:swahili_layerwise} and Figure \ref{fig:hindi_layerwise} present results for Swahili and Hindi respectively.


\begin{figure}
\includegraphics[width=\columnwidth]{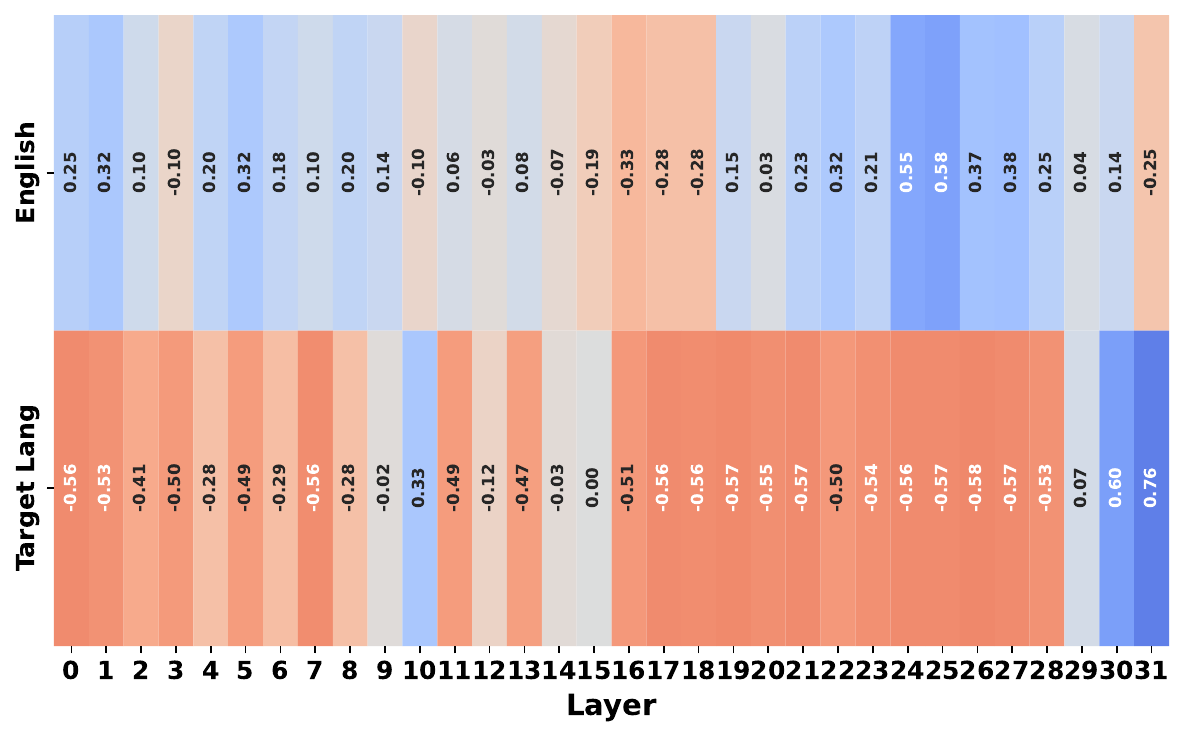}
  \caption{Correlation of layer-wise TL ratio and English ratio to TL F1 score. Warm colors (reds/oranges) denote negative correlations, while cool colors (blues) denote positive correlations.}
  \label{fig:correlation_heatmap}
\end{figure}


\end{document}